\documentclass[sn-mathphys]{sn-jnl}% Math and Physical Sciences Reference Style
\theoremstyle{thmstyleone}%
\theoremstyle{thmstyletwo}%

\theoremstyle{thmstylethree}%

\begin{document}

\title[A Neighbor-based Approach to Pitch Ownership Models in Soccer]{A Neighbor-based Approach to Pitch Ownership Models in Soccer}


\author*[1,2]{\fnm{Tiago} \sur{Mendes-Neves}}\email{tiago.neves@fe.up.pt}

\author[3]{\fnm{Luís} \sur{Meireles}}\email{luis.meireles@nordensa.com}

\author[1,2]{\fnm{João} \sur{Mendes-Moreira}}\email{jmoreira@fe.up.pt}

\affil*[1]{\orgdiv{Faculdade de Engenharia}, \orgname{Universidade do Porto}, \orgaddress{\city{Porto}, \country{Portugal}}}

\affil[2]{\orgdiv{LIAAD}, \orgname{INESC TEC}, \orgaddress{\city{Porto}, \country{Portugal}}}

\affil[3]{\orgname{Nordensa Football}, \orgaddress{\city{Cluj}, \country{Romania}}}


\abstract{
Pitch ownership models allow many types of analysis in soccer and provide valuable assistance to tactical analysts in understanding the game's dynamics. The novelty they provide over event-based analysis is that tracking data incorporates context that event-based data does not possess, like player positioning. 
This paper proposes a novel approach to building pitch ownership models in soccer games using the K-Nearest Neighbors (KNN) algorithm. Our approach provides a fast inference mechanism that can model different approaches to pitch control using the same algorithm. Despite its flexibility, it uses only three hyperparameters to tune the model, facilitating the tuning process for different player skill levels. The flexibility of the approach allows for the emulation of different methods available in the literature by adjusting a small number of parameters, including adjusting for different levels of uncertainty.
In summary, the proposed model provides a new and more flexible strategy for building pitch ownership models, extending beyond just replicating existing algorithms, and can provide valuable insights for tactical analysts and open up new avenues for future research. We thoroughly visualize several examples demonstrating the presented models' strengths and weaknesses. The code is available at github.com/nvsclub/KNNPitchControl.
}




\maketitle

\section{Introduction}\label{sec:introduction}

Using tracking data in tactical analysis is a relatively new phenomenon for soccer. The reason is simple: tracking data was only available through wearable tracking devices, which the International Football Association Board allowed in 2015. At the same time, the accuracy of object detection algorithms was improving substantially, enabling optical tracking data from video broadcasts. Although GPS tracking data is much more reliable and accurate, there is no central repository where teams share this data. Therefore, most teams only have access to the tracking data of their team. With optical tracking data, it is easier to access completer data sets, with information from both teams and the ball.

Traditional statistical analysis in soccer, such as goals and possession, or more advanced statistical techniques, like expected goals, can be limited in their ability to fully capture important aspects of the opposition's style of play. Tracking data gives a more comprehensive picture of a player's performance, allowing for more accurate evaluation and tactical decision-making.

The most immediate use case for tracking data is measuring players' physical performance. It allows a better understanding of preventing injuries and getting the best out of players. With the increased scale of data, new areas started taking advantage of tracking data. Tactical analysis using pitch ownership models is one of them.

\begin{figure}[h]
\centering
\includegraphics[width=\linewidth]{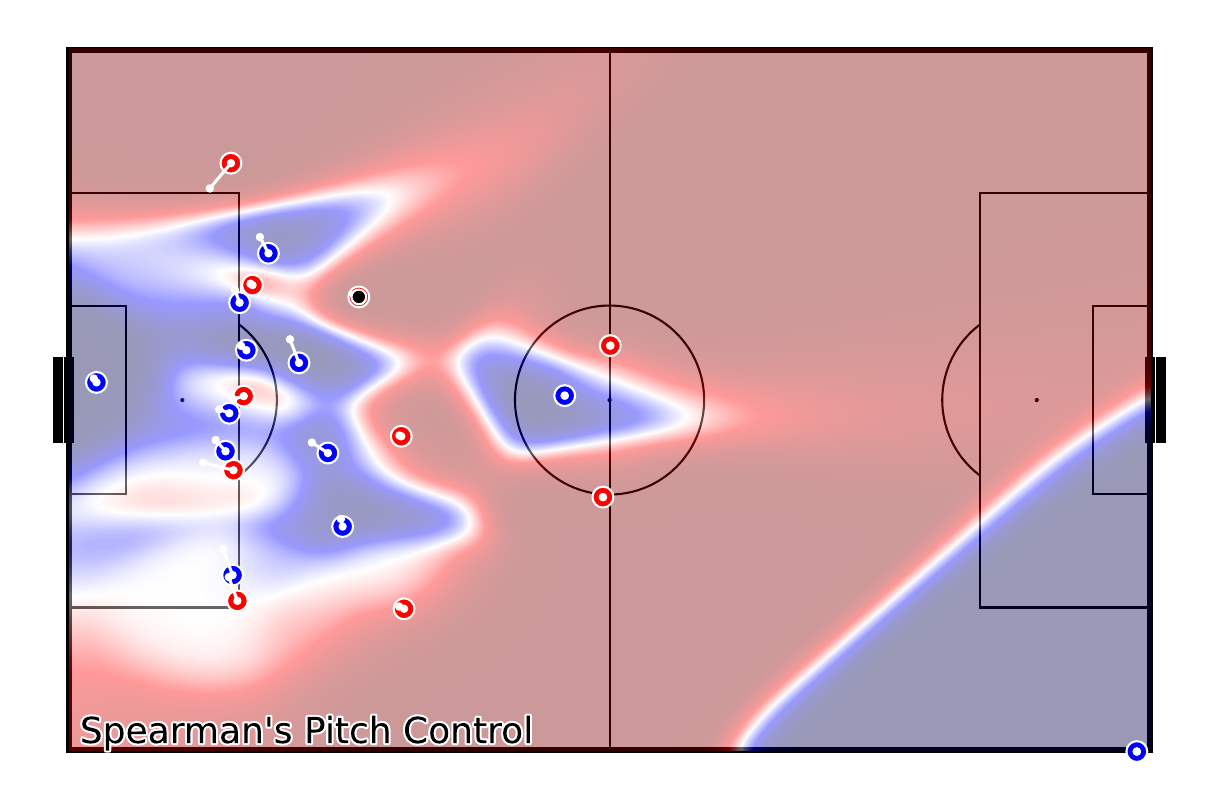}
\caption{A visualization of the pitch control surface as described by Spearman - implementation by Laurie Shaw. The blue team dominates blue areas. The red team dominates red areas. Light-colored areas have higher uncertainty over who controls the area. White areas indicate that there is complete uncertainty over who controls this area.}\label{fig:spearman_example_default}
\end{figure}

Pitch ownership is a proxy measure for which a team "controls" a specific area of the pitch. The definition of "control" varies according to the methodology used. Voronoi diagrams define control as the space in which a team's player is closest. Spearman \cite{spearman_beyond_2018} describes pitch control focusing on possession: "the probability that the home team will end up with possession of the ball if it were at the location, x." Spearman's version of pitch control can be viewed in Figure \ref{fig:spearman_example_default}. For Fernandez et al. \cite{fernandez_wide_2018}, pitch control is "the degree or probability of control that a given player (or team) has on any specific point in the available playing area."

Although definitions might vary slightly, they all have the same goal: understanding where are areas of strength and weakness to provide an advantage for the team. Pitch ownership models allow coaches to make informed decisions about the team's weaknesses. It provides players with visual representations of their mistakes, which makes it much easier to understand tactical decisions and judge a player's acumen.

In our paper, we adopt Spearman’s definition of control. The goal is to measure the likelihood of a player controlling the ball in a specific space. Furthermore, we introduce a few parameters that allow our model to cover a broad spectrum of assumptions.

We use the K-Nearest Neighbors (KNN) algorithm to build our pitch ownership model. A 1-Nearest Neighbor (1NN) of the tracking data provides information about which player is closest to a position in the pitch, which approximates the Voronoi diagram \cite{clark_distance_1954}. This property of the KNN algorithms makes it suitable for implementing pitch control. We build on top of this definition to combine multiple Voronoi diagrams that approximate other approaches in the literature and enable the construction of novel types of models. 

To introduce uncertainty into our models, we present two parameters. \textbf{Distance decay} models control based on players' proximity to a certain point, while \textbf{smoothing} applies moving averages to smooth the control matrix.
Modeling uncertainty is a crucial step for pitch ownership models. For example, lower leagues will have a higher uncertainty because players take more time to control the ball, and we can use these parameters to tune the resulting models. 

The main contribution of this paper is a novel strategy to build pitch ownership models. Our approach is more flexible than Voronoi’s, Spearman’s, and Fernandez’s, requiring fewer parameters and assumptions. Due to the lack of extensive datasets, these models cannot be valued quantitatively. Therefore we thoroughly review examples that describe the strengths and weaknesses of each of the presented models.

This paper is organized as follows:
\begin{itemize}
    \item Section \ref{sec:pitchownershipmodels} defines pitch ownership models and presents the different approaches to obtaining them.
    \item Section \ref{sec:knnpitchcontrol} presents our approach that uses KNN to model pitch control, including the different parameters available.
    \item Section \ref{sec:methodology} describes the methods used to generate the different pitch control models. Section \ref{sec:results} presents a sample of the results obtained with this methodology.
    \item Section \ref{sec:discussion} discusses the results obtained and \ref{sec:conclusion} presents the concluding remarks of this paper.
\end{itemize}

\section{Pitch Ownership Model}\label{sec:pitchownershipmodels}

Tracking data possesses many use cases. Llana et al. \citep{llana_is_2022} built a framework that contains most of the novel techniques: (1) finding high-intensity runs, (2) calculating team lines, (3) calculating the shape of the defensive block, (4) identification of formation and player roles, (5) pass likelihood, (6) expected possession value and (7) pitch control. These techniques provide the building blocks to build statistical analysis tools in soccer. Most statistical tools are currently built on event data, which is data collected from on-ball events such as passes, shots, and tackles, among others \cite{decroos_actions_2019, singh_introducing_2019}. 

Pitch control is a pitch ownership model. Taki et al. \citep{taki_development_1996} published the first resemblance of pitch control, where he defined dominant regions. This work accounted for the most important variables: player position, speed, and direction. Control was defined as "as a region where the player can arrive at earlier than all of the others". The goal was to evaluate teamwork by observing how player movement and runs affected the regions dominated by each team.

Although Voronoi diagrams existed for over a century, \cite{voronoy_nouvelles_1908}, they have only recently been used for tactical analysis. Fonseca et al. \cite{fonseca_spatial_2012} introduced using Voronoi diagrams to explore the spacial dynamics of players in futsal, and validate important insights like attackers usually controlling larger and less important regions versus the defenders.

Spearman defines a physics-based pitch ownership model, which uses the concept of electric potential fields and applies it to tracking data. The electric potential fields quantify a particular particle's force across space. Similar to a soccer game, particles belong to two\footnotemark[1] "teams": negative charge and positive charge. Therefore, by attributing to each team a specific charge (e.g., the home team has a charge equal to 1, away team equal to -1), we can see players as charges in space. Equation \ref{eq:spearman_pc} describes how to calculate Spearman's version of pitch control, where $i$ is a player, $t_i$ represents the time it takes for a player to reach the ball, $l_i$ is the player's team, and $\beta$ is a parameter to tune how important it is to the player to be the first to reach the ball.

\footnotetext[1]{We ignore neutral charges because they do not affect the result.}

\begin{equation}
PCF(t_i,l_i) = \frac{\sum_i{l_it_i^{-\beta}}}{\sum_i{t_i^{-\beta}}}
\label{eq:spearman_pc}
\end{equation}

Equation \ref{eq:spearman_pc} only simplifies the actual necessary calculations. For instance, the value of $t_i$ requires calculating how much time is expected for the player to reach the ball. This parameter requires specific assumptions about a player's speed and acceleration, either a constant value for all players or individually set. Furthermore, we require data to estimate the parameter $\beta$. Spearman \cite{spearman_beyond_2018} describes a procedure to estimate $\beta$, to which he estimates $\beta \in [2.3, 2.7]$. However, this parameter can change substantially with a league's quality and playing intensity. Note that we get a Voronoi diagram for $\beta \to +\inf$.

Fernandez et al. \cite{fernandez_wide_2018} define an approach to pitch ownership models using bivariate normal distributions to model a player's influence area.
The distribution's mean and covariance matrix of each player accounts for the player's location and speed. By summing the distributions of individual players, Fernandez et al. \cite{fernandez_wide_2018} calculate pitch control for the whole pitch.

Pitch control has many use cases. Spearman \cite{spearman_beyond_2018} suggests several video analysis applications such as detecting players' defensive positioning mistakes, missed offensive opportunities, evaluating passing decisions, player positioning quality, and quantifying the effect of player positioning. More recently, Fernandez et al. \cite{fernandez_wide_2018} included pitch control on a framework to measure space creation in soccer. Martens et al. \cite{martens_space_2021} devise space generation metrics for game analysis. Pitch control can detect key moments in a match, such as counter-attacks, set pieces, and defensive/offensive transitions.

The drawdown of the use of tracking data is the lack of quality in some datasets. Some datasets are incomplete, which is common in tracking data obtained through broadcast video. There is noise in the data due to occlusions, changes in image quality, and similar problems. Furthermore, players can be wrongly identified. Some methods require calibration, which can be performed poorly, and inconsistent providers use different standards to obtain tracking data. Nevertheless, some of the problems identified are currently being researched in computer vision, with significant advances in correcting the quality of tracking datasets \cite{omidshafiei_multiagent_2022}.

\section{KNN-based Pitch Control}\label{sec:knnpitchcontrol}
\subsection{Concepts}
Our proposal uses three key concepts:

\begin{itemize}
    \item Using the KNN algorithm to approximate Voronoi diagrams.
    \item Introduce uncertainty with the KNN’s distance.
    \item Introduce player interaction uncertainty with smoothing via moving averages.
\end{itemize}

\newpage

\subsubsection{Approximating Voronoi diagrams with KNN}

The relationship between the KNN algorithm and Voronoi diagrams is important because KNN closely maps the underlying Voronoi diagram, enabling us to combine multiple variants of the Voronoi diagram.

\begin{figure}[h]
\centering
\includegraphics[width=\linewidth]{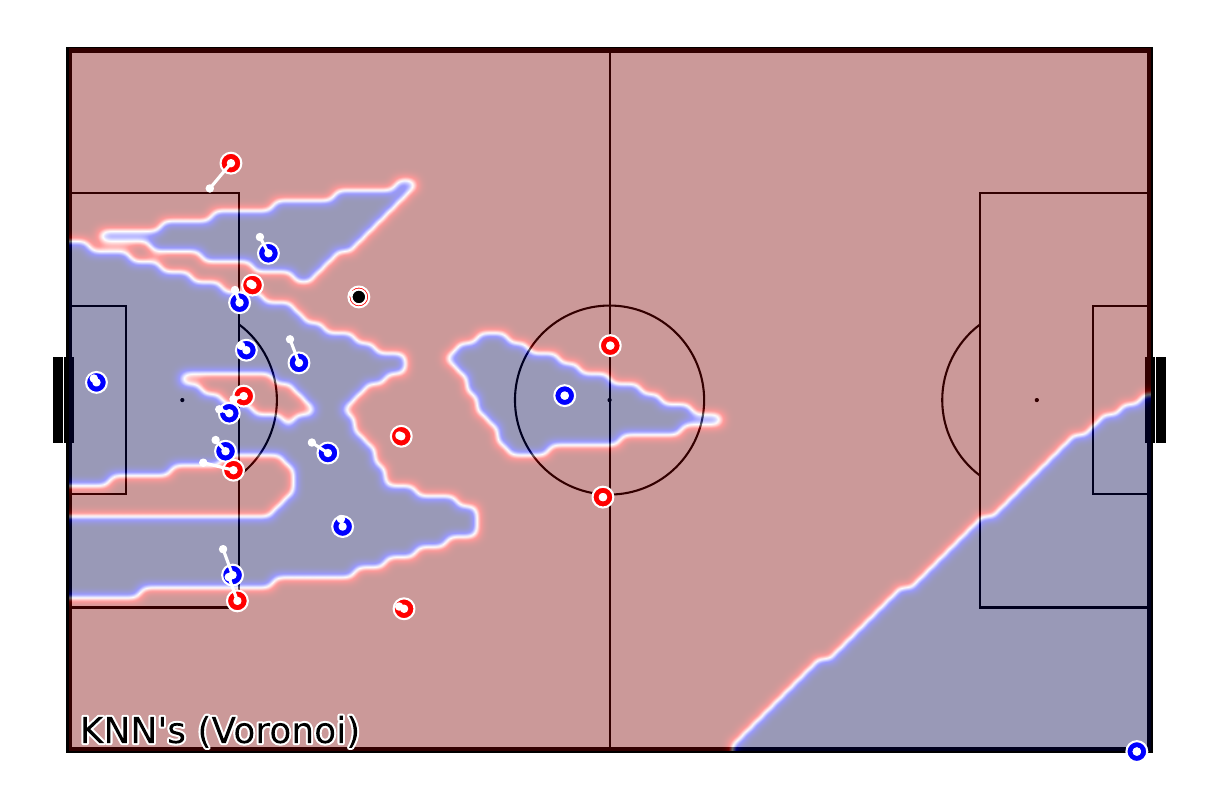}
\caption{An approximation of the Voronoi diagram using the KNN algorithm, for a resolution of $1m^2$.}\label{fig:voronoi_default}
\end{figure}

The K-Nearest Neighbors (KNN) algorithm is a machine learning technique for clustering and classification tasks. In the context of soccer analytics, the KNN algorithm approximates Voronoi diagrams, which define areas of the pitch closest to each player, as observed in Figure \ref{fig:voronoi_default}.

The basic idea behind using KNN to approximate Voronoi diagrams is to identify the player closest to a given point on the pitch. This procedure can be done by calculating the Euclidean distance between each player and the point in question and selecting the player with the smallest distance. Repeating this process for every point on the pitch makes it possible to construct a Voronoi diagram that approximates the pitch areas closest to each player.

The KNN algorithm approximates Voronoi diagrams as the resolution of the grid tends to infinity. For resolutions of $1m^2$, it is an accurate and computationally efficient way to model pitch control. Furthermore, combining multiple Voronoi diagrams that approximate different approaches in the literature makes building more flexible and precise pitch ownership models possible. This approach can also enable the construction of novel models that are impossible with traditional Voronoi diagrams or other methods.

\subsubsection{Introducing uncertainty with KNN's distance}

\begin{figure}[h]
\centering
\includegraphics[width=\linewidth]{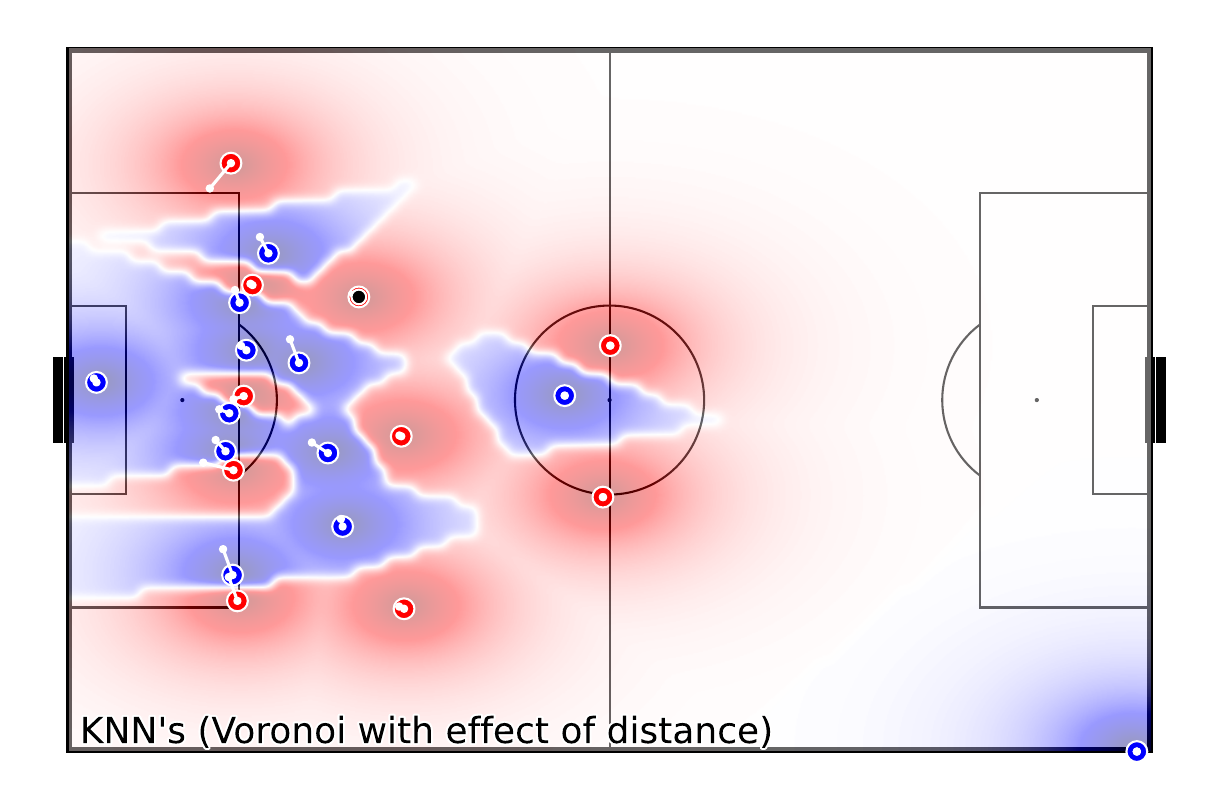}
\caption{The effect of modeling distance on the Voronoi diagram.}\label{fig:voronoi_with_distance}
\end{figure}

One advantage of building Voronoi diagrams with KNN is that we can use the distances calculated to introduce uncertainty in areas of the pitch that are further away from the player. The larger the distance a player has to cover, the more significant the impact of unknown player factors such as reaction speed and velocity. These unknown factors influence a player's control over a distant position. This parameter allows the model to account that players further away from the ball are less likely to control it than closer players. Equation \ref{eq:distance_effect} defines this procedure mathematically, where $C$ is a computed control matrix, $x, y$ the coordinates of the points in the matrix, $\xi$ is a user-defined parameter that tunes how fast control decays with distance, and $s$ is the distance between the player and the $x,y$ coordinate being calculated. The larger the value of $\xi$, the larger the uncertainty introduced by the effect of distance, as observed in Figure \ref{fig:voronoi_with_distance}.

\begin{equation}
Distance\ C (x, y) = C(x,y) * \xi ^ {1-\frac{s}{40}} 
\label{eq:distance_effect}
\end{equation}

\newpage

\subsubsection{Introducing uncertainty with smoothing} \label{subsec:smoothing}
The Voronoi diagram classifies the region of a player as being controlled by his team. However, in soccer, there is high uncertainty in regions where players might dispute possession. Therefore, it is important to have a way to add uncertainty in these places. We likely only need to add a little uncertainty for top-tier teams since players generally have high technical ability and situational awareness. However, in lower tiers, players have lower quality; therefore, we need to add more uncertainty since they are more likely to make mistakes.

\begin{figure}[h]
\centering
\includegraphics[width=\linewidth]{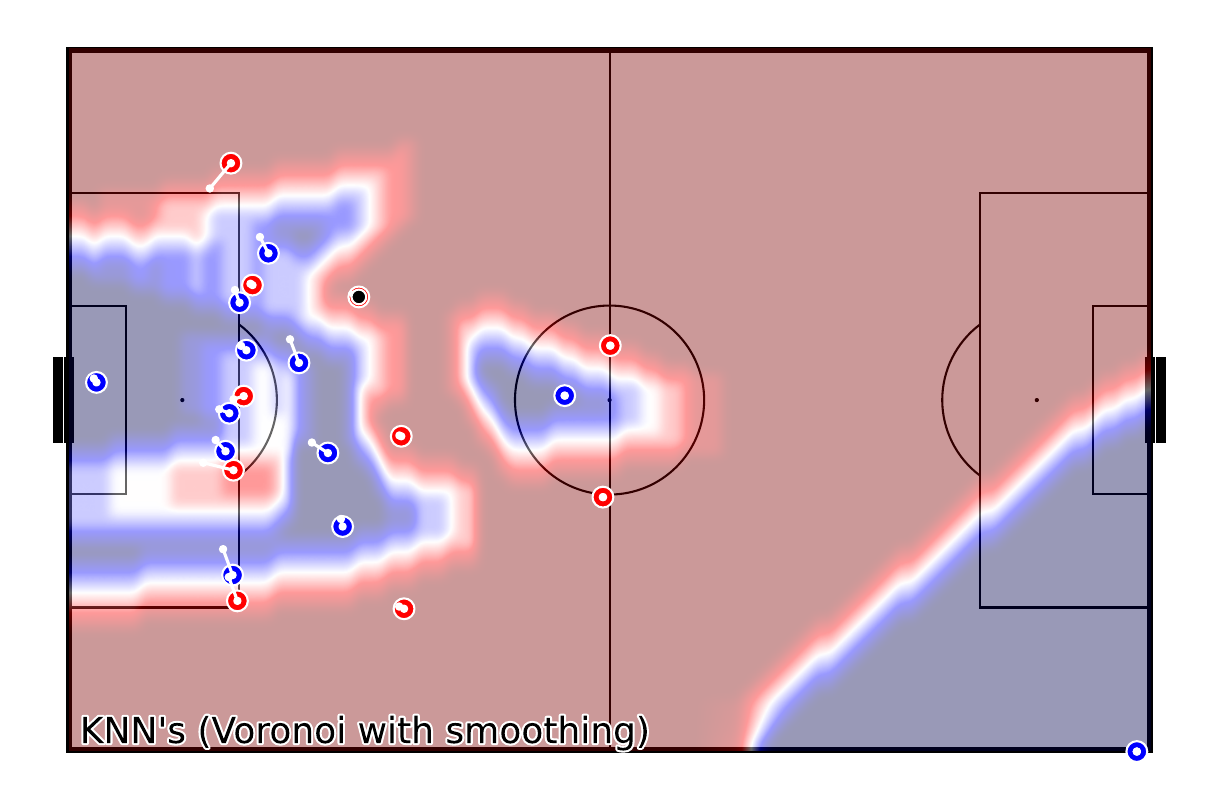}
\caption{The areas of uncertainty obtained when smoothing are visible in the lighter spaces.}\label{fig:voronoi_smoothing}
\end{figure}

In our work, we use a double moving average of an interval $\tau$ across both axis in the results grid. We apply the operation mathematically defined in Equation \ref{eq:smoothing} two times sequentially on each axis. $C$ is a computed control matrix, $x, y$ is the signal in which we apply the moving average, and $\tau$ is the user-defined window. The smoothing effect is visible in Figure \ref{fig:voronoi_smoothing}.

\begin{equation}
Smooth\ C (x, i) = \frac{\sum_{j=i-\frac{\tau}{2}}^{i+\frac{\tau}{2}} C(x, j) }{\tau²},\ or\ Smooth\ C (i, y) = \frac{\sum_{j=i-\frac{\tau}{2}}^{i+\frac{\tau}{2}} C(j, y) }{\tau²} 
\label{eq:smoothing}
\end{equation}

\newpage

\subsection{Combining the concepts}

When instantiating our method, the user defines three parameters: the $lags$, $\xi$, and $\tau$. Internally, our method follows three steps: generating the Voronoi diagrams using KNN, adding the distance factor, and smoothing the gradient. Note that the user can skip the latter two.

The first step uses the lags parameter, where the user gives a list of $lags$ to build the Voronoi diagrams. The goal of a lag is to extrapolate current player speed and incorporate this information in the pitch control diagram.

For each $lag\ \eta$, the method will generate a Voronoi diagram with the position of each player calculated according to Equation \ref{eq:combinations}, where $p$ is the position calculated in the frame $i$ and $x, y$ are the coordinates of the players’ positions.

\begin{equation}
p_i = (x_i + \eta dx_i, y_i + \eta dy_i), where\ dx=x_i-x_{i-1}, and\ dy=y_i-y_{i-1}
\label{eq:combinations}
\end{equation}

After calculating each Voronoi diagram, we calculate the effect of distance through Equation \ref{eq:distance_effect}. Note that if the user gives no value to $\xi$ or sets $\xi=1$, we ignore this step, meaning that the user does not want the effect of distance to impact the final gradient.

Having calculated every Voronoi diagram and the effect of distance, we sum all Voronoi diagrams together. The last step is to smooth the gradient using the procedure described in Section \ref{subsec:smoothing}.

\section{Methodology}\label{sec:methodology}

The data used in this work is available in the Friends of Tracking repository\footnotemark[2]. Specifically, our dataset contains the plays of 20 goals, 19 scored by Liverpool in 2019 and 1 by Real Madrid in 2012.

\footnotetext[2]{https://github.com/Friends-of-Tracking-Data-FoTD/Last-Row}

To produce the results presented in Section \ref{sec:results}, we use the following models:

\begin{itemize}
    \item KNN (Voronoi) - Our approach with parameters $\eta=[0]$, $\xi=None$, and $\tau=None$, which produces a default Voronoi diagram as a baseline.
    \item KNN (Spearman) - Our approach with parameters $\eta=[0, 10, 25]$, $\xi=None$, and $\tau=6$ produces a visualization with principles similar to Spearman’s approach.
    \item KNN (Fernandez) - Our approach with parameters $\eta=[0, 10, 25]$, $\xi=350$, and $\tau=6$ produces a visualization with principles similar to Fernandez’s approach.
    \item Spearman Pitch Control - An implementation of Spearman’s pitch control by Laurie Shaw\footnotemark[3]. We use the default parameters but treat goalkeepers as regular players due to restrictions in our dataset.
    \item Fernandez Pitch Control - An implementation of Fernandez pitch control by Will Thomson\footnotemark[4].
\end{itemize}

\footnotetext[3]{https://github.com/Friends-of-Tracking-Data-FoTD/LaurieOnTracking}
\footnotetext[4]{https://github.com/anenglishgoat/Metrica-pitch-control}

We do not provide information about inference time due to substantial differences in the implementations. However, note that the tested algorithms have implementations that can easily produce inferences in real-time, e.g., Will Thomson’s version of Spearman’s pitch control can be generated in under 30 seconds for an average game of soccer.

The implementation code is available on GitHub\footnotemark[5].
\footnotetext[5]{https://github.com/nvsclub/KNNPitchControl}

\section{Results}\label{sec:results}

\begin{figure}[h]
    \centering
    \includegraphics[height=3.5cm,width=0.45\linewidth]{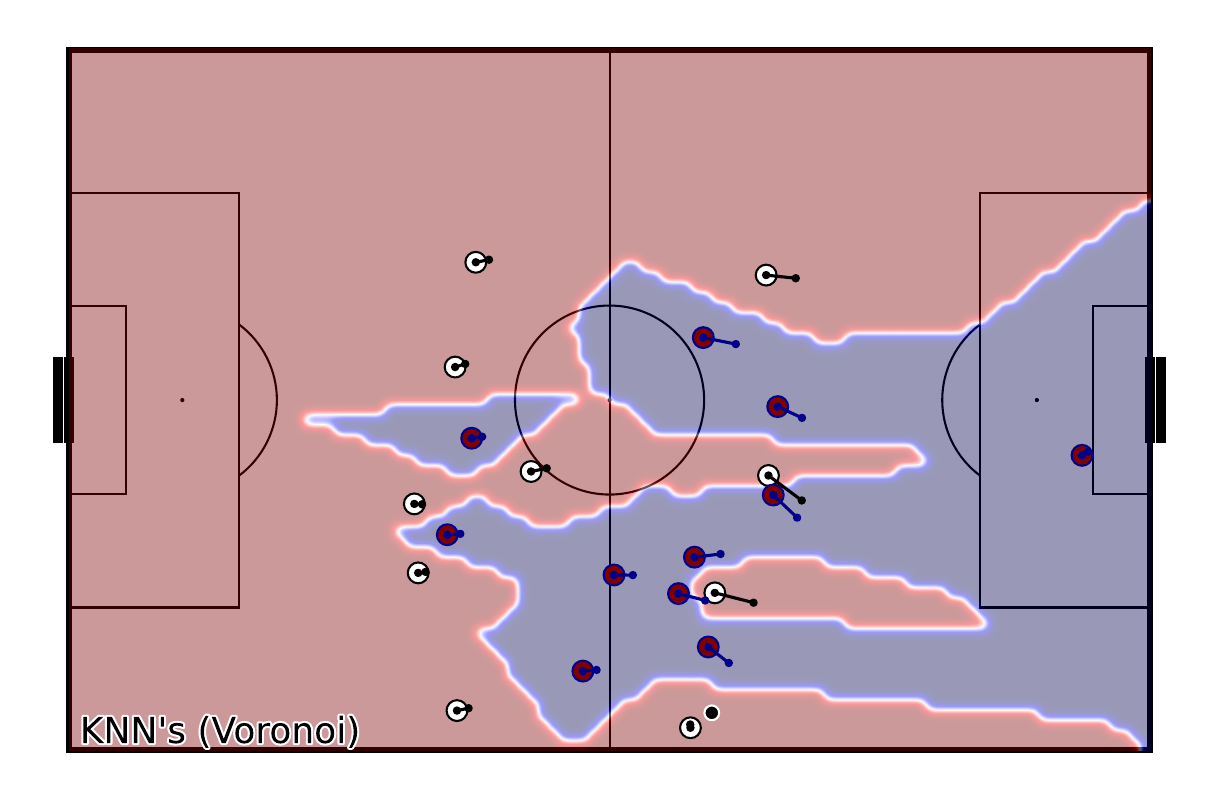}\\
    \includegraphics[height=3.5cm,width=0.45\linewidth]{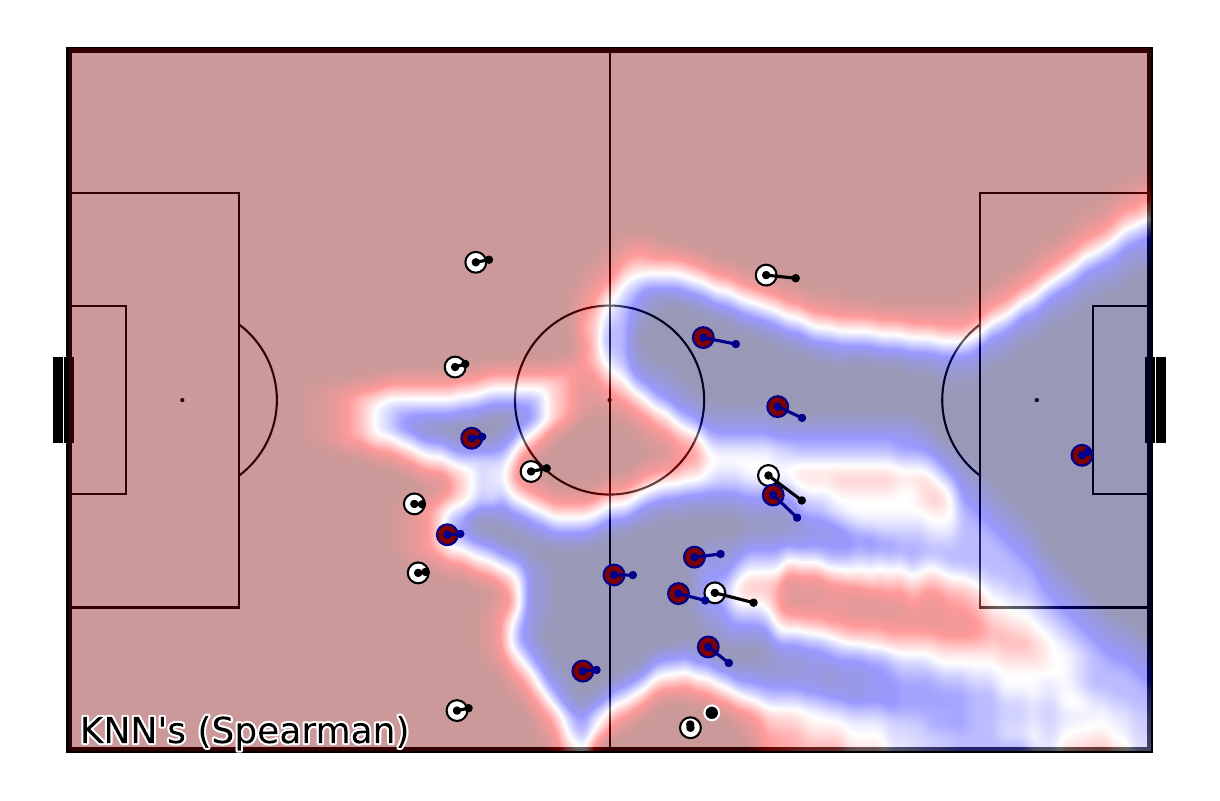}
    \includegraphics[height=3.5cm,width=0.45\linewidth]{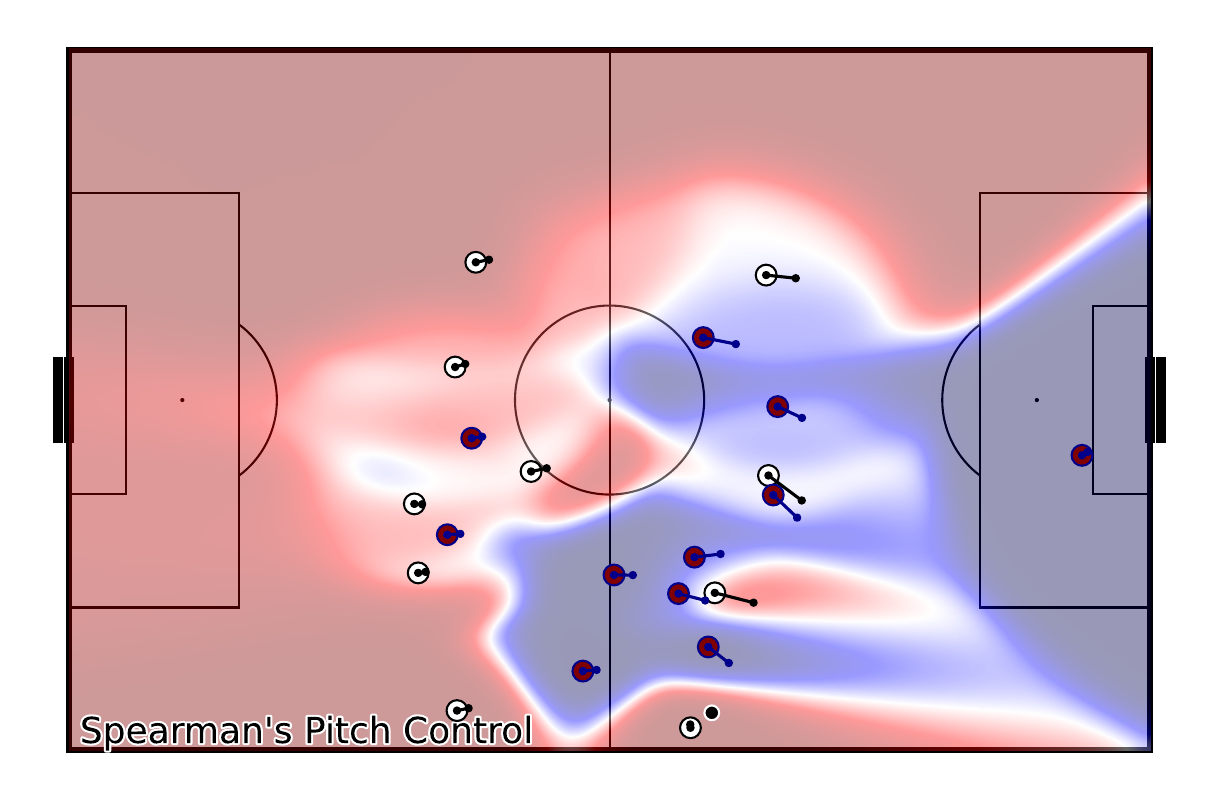}\\
    \includegraphics[height=3.5cm,width=0.45\linewidth]{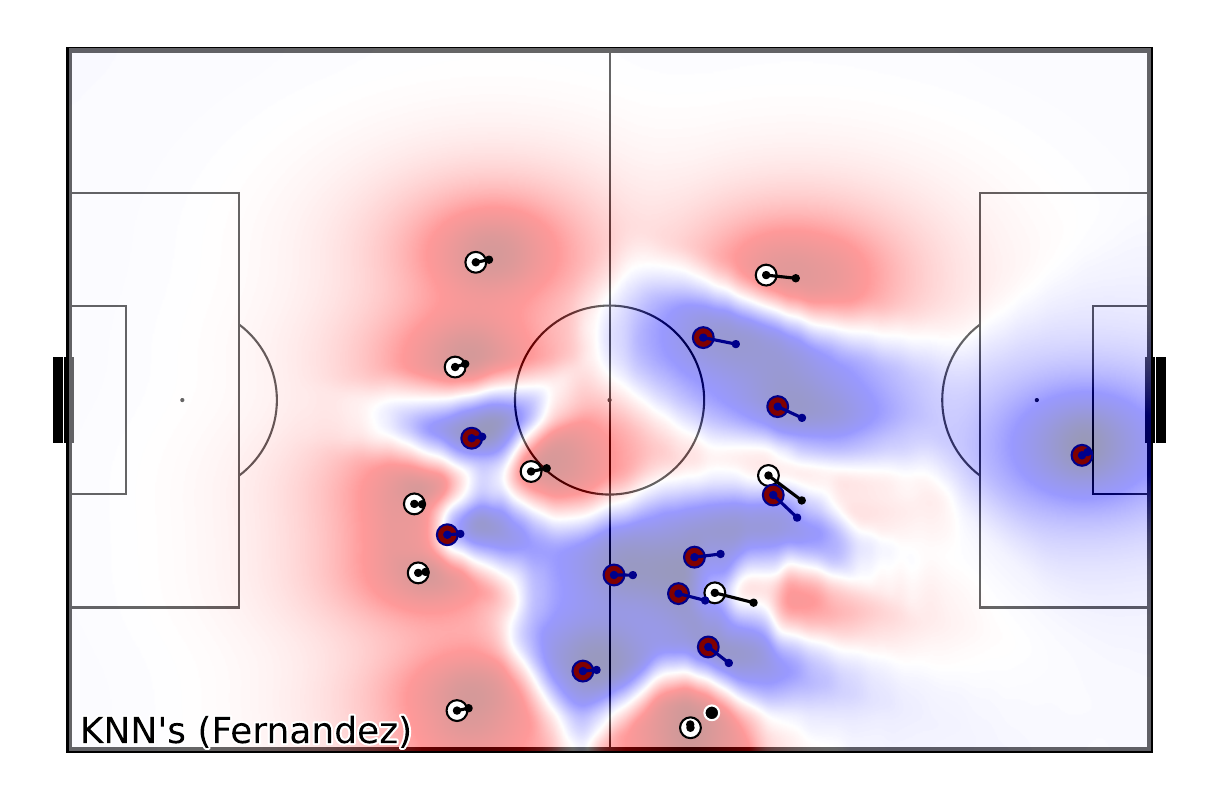}
    \includegraphics[height=3.5cm,width=0.45\linewidth]{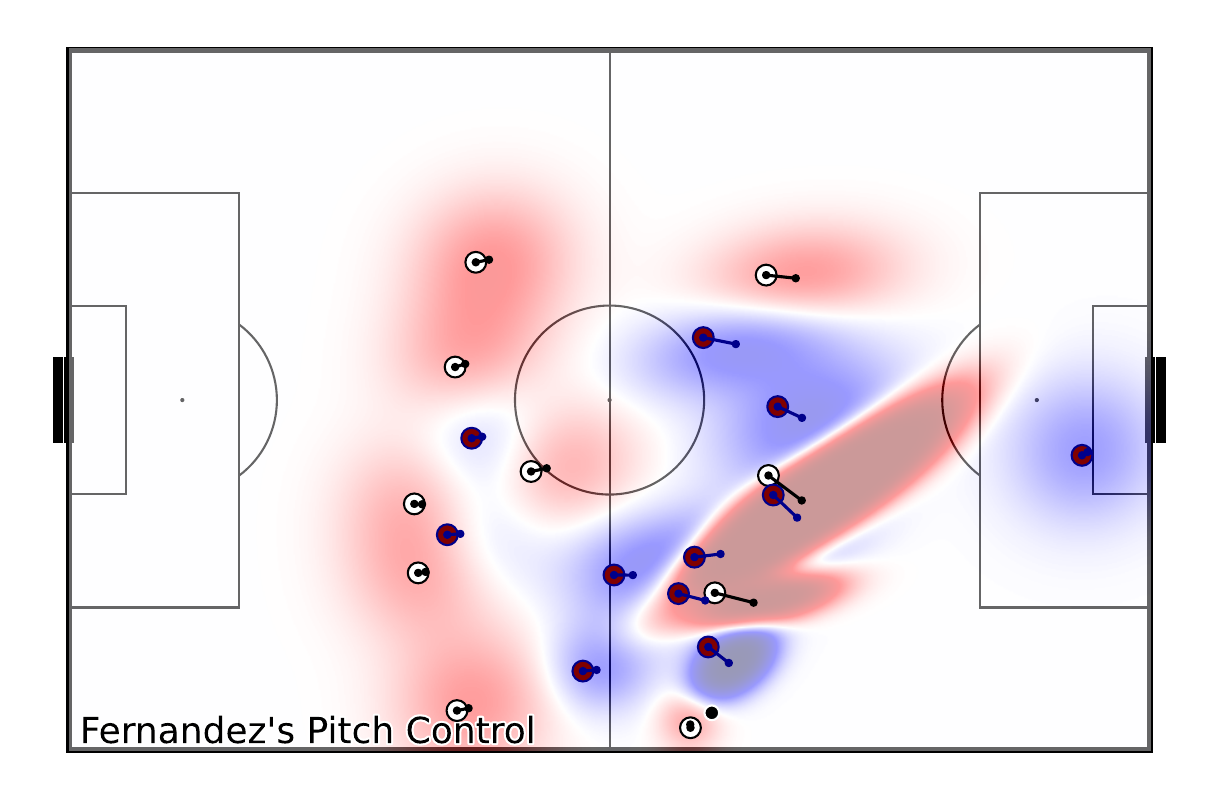}
    \caption{The results obtained with KNN-based Pitch Control compared with the original approaches, according to the parameters set in Section \ref{sec:methodology}. On the left, we present our approaches; on the right, we present the similar approaches available in the literature. The results we present are from the play Barcelona 1 - [2] Real Madrid on the 21st of April 2012, frame 132.}
    \label{fig:results}
\end{figure}

Figure \ref{fig:results} presents a visualization of the several methods to generate pitch control gradients for comparison.

\section{Discussion}\label{sec:discussion}
Voronoi diagrams are the most straightforward method to generate pitch control. However, Voronoi diagrams have their drawdowns. The most important drawdown is that it does not account for player speed. The static assumptions of a Voronoi diagram ignore the high dynamism of a soccer match. Compared with other diagrams, it is clear that the strategies adopted to account for player speed provide better and more precise insights. 

Incorporating player speed was precisely one of our main goals. We observe the influence of the parameter $lags$ in Figure \ref{fig:results}. The ability to extrapolate current player speed improves the insights' accuracy. For example, the space behind the blue team's defensive line more accurately reflects the current dynamics in the game.

Our approach contains some areas that disagree with current approaches defined in the literature. While our approximation of Spearman's approach is reasonably accurate, it disagrees in some areas of uncertainty. This is because our approach does not account for the time it takes the ball to travel to another position. We believe that not accounting for the ball's travel time is not an inherently bad thing. From the point of view of tactical analysis, it makes sense not to make hard assumptions about the ball position since it allows analysts to assess alternative hypotheses. For example, judging the value added from a pass by whether it moved the ball to a position that allowed a player to explore a controlled area of importance.

Regarding Fernandez's approach, our model's approximation disagrees substantially. First, the way Fernandez models control behind the blue team's line of defense is very different from all other approaches. Initially, we suspected that the orientation of the distribution was calculated in the wrong direction, but we concluded this was not the case due to the correct behavior in all other frames. The second disagreement is that, according to the implementation of Fernandez's pitch control used, the results did not have a fixed range. This leads to inconsistencies when visualizing this approach.

One problem that no approach solves is body blocking. When a player is in front of an opposing player's path, he can significantly disturb the attacking player's trajectory, affecting how much control the attacking player has. Although this is a minor detail, it is important to take notice when analyzing a play using any of the methods.

Due to the nature of how we calculate our pitch control, our approach looks less smooth than the other approaches. One option to increase the smoothness is to increase the density of the grid with which we calculate the pitch control. As density increases, the inference time will also increase. The density used in the plots is 105x68 points, which relates to the typical size of a football pitch (105m x 68m).

\section{Conclusion}\label{sec:conclusion}

In conclusion, our paper presents a novel and flexible approach to building pitch ownership models in soccer games using the KNN algorithm. Pitch control has become an important method of analysis in soccer, and our model provides valuable assistance to tactical analysts in understanding the game's dynamics. The assistance it provides is very valuable since it exposes some relations that are not evident by watching the video footage.
Our approach allows us to emulate different methods available in the literature by adjusting a small number of parameters. Our model accounts for player speed and different types of uncertainty. We observed that the insights provided by our model could differ substantially from other approaches, highlighting the importance of having multiple methods to analyze pitch control.
Furthermore, our proposed approach raises the question of how other machine learning algorithms can generate interesting insights from tracking data, particularly calculating areas of control for each team. Future work could focus on developing a systematic and quantitative way of measuring the quality of pitch control models. Another relevant open question is accounting for body blocking. 
Despite the lack of extensive datasets, we thoroughly reviewed examples demonstrating the presented models' strengths and weaknesses. Our approach requires fewer parameters and assumptions, making it a valuable tool for tactical analysis.
In summary, our model provides a new, more flexible strategy for building pitch ownership models, extending beyond replicating existing algorithms. It can provide tactical analysts valuable insights and open new avenues for future research.

\bmhead{Acknowledgments}

This work is financed by National Funds through the Portuguese funding agency, FCT - Fundação para a Ciência e a Tecnologia, within project UIDB/50014/2020.

\bibliography{sn-bibliography}% common bib file


\end{document}